\documentclass[journal,lettersize,twoside]{IEEEtran}

\usepackage{graphicx}
\usepackage{cite}
\usepackage{amsmath}
\usepackage{amssymb}
\usepackage[ruled, lined, longend, linesnumbered]{algorithm2e}
\usepackage{array}
\usepackage{float}
\usepackage{multirow}
\usepackage[utf8]{inputenc}
\usepackage[hidelinks]{hyperref}
\usepackage{framed}
\usepackage{picinpar}
\usepackage{url}
\usepackage{flushend}
\usepackage{colortbl}
\usepackage{soul}
\usepackage{pifont}
\usepackage{color}
\usepackage{alltt}
\usepackage{enumerate}
\usepackage{siunitx}
\usepackage{breakurl}
\usepackage{epstopdf}
\usepackage{pbox}

\newtheorem{assumption}{Assumption}
\newtheorem{proof}{Proof}
\newtheorem{prop}{Proposition}

\begin{document}

\author{Ehsan Hallaji,~\IEEEmembership{Graduate Student Member,~IEEE,}
        Roozbeh Razavi-Far,~\IEEEmembership{Senior Member,~IEEE,}
        Meng Wang,~\IEEEmembership{Member,~IEEE,}
        Mehrdad Saif,~\IEEEmembership{Senior Member,~IEEE,}
        and Bruce Fardanesh,~\IEEEmembership{Fellow,~IEEE}
\thanks{Ehsan Hallaji and Mehrdad Saif are with the Department of Electrical
and Computer Engineering, University of Windsor, Windsor, ON N9B 3P4,
Canada (e-mail: hallaji@uwindsor.ca; msaif@uwindsor.ca).}
\thanks{Roozbeh Razavi-Far is with the Faculty of Computer Science, University of New Brunswick, Fredericton, NB E3B 5A3, Canada and also with the Department of Electrical and Computer
Engineering, University of Windsor, Windsor, ON N9B 3P4, Canada (e-mail: roozbeh.razavi-far@unb.ca).}
\thanks{Meng Wang is with the Department of Electrical, Computer, and Systems
Engineering, Rensselaer Polytechnic Institute, Troy, NY 12180 USA (e-mail:
wangm7@rpi.edu).}
\thanks{Bruce Fardanesh is with the Department of
Research and Technology Development, New York Power Authority,
White Plains, NY 10601 USA (e-mail: bruce.fardanesh@nypa.gov).}}

\title{A Stream Learning Approach for Real-Time Identification of False Data Injection Attacks in Cyber-Physical Power Systems}

\markboth{IEEE Transactions on Information Forensics and Security}
{HALLAJI \MakeLowercase{\textit{et al.}}: A Stream Learning Approach for Real-Time Identification of FDI in Cyber-Physical Power Systems}

\maketitle

\begin{abstract}
This paper presents a novel data-driven framework to aid in system state estimation when the power system is under unobservable false data injection attacks. The proposed framework dynamically detects and classifies false data injection attacks. Then, it retrieves the control signal using the acquired information. This process is accomplished in three main modules, with novel designs, for detection, classification, and control signal retrieval. The detection module monitors historical changes in phasor measurements and captures any deviation pattern caused by an attack on a complex plane. This approach can help to reveal characteristics of the attacks including the direction, magnitude, and ratio of the injected false data. Using this information, the signal retrieval module can easily recover the original control signal and remove the injected false data. Further information regarding the attack type can be obtained through the classifier module. The proposed ensemble learner is compatible with harsh learning conditions including the lack of labeled data, concept drift, concept evolution, recurring classes, and independence from external updates. The proposed novel classifier can dynamically learn from data and classify attacks under all these harsh learning conditions. The introduced framework is evaluated w.r.t. real-world data captured from the Central New York Power System. The obtained results indicate the efficacy and stability of the proposed framework.
\end{abstract}


\section{Introduction}
\IEEEPARstart{T}{he} advancement of information technology has improved smart grids in many aspects such as reliability and efficiency. Nonetheless, security concerns arise, as the state estimation of power systems becomes more dependent on the software components of smart grids \cite{9207760, 8660426}. Intruders can interfere with the operation of the control system by launching a False Data Injection (FDI) attack. Having enough information about the system configuration, a hacker can manually corrupt the obtained phasor measurements and prevent the control system to correctly estimate the system state. This can, in turn, cause a malfunction that entails potentially catastrophic consequences \cite{7867825, 7433442}.

Although various approaches can be used to enhance the security of the power networks, they may not ideally replace a robust Intrusion Detection System (IDS). For instance, encryption does not always guarantee the safety of the system, and grid hardening may not be applicable to large-scale power systems \cite{7792204,7529082}. Although various approaches are available for eliminating security issues in power systems, not all of them can cope with the unobservable FDI attacks. Such attacks manipulate multiple phasor measurements and deceive the control system by bypassing the bad data detectors. These FDI attacks are called unobservable since the system is no longer observable if the affected measurements are removed \cite{7529082, 7792204}. Perhaps, unobservable attacks can be considered as the most challenging type of cyber data attacks \cite{7529082}. System state estimation under unobservable FDI attacks requires a robust IDS to ensure the safe operation of the control system. Despite numerous efforts dedicated to addressing this problem, finding a comprehensive design for an IDS is still very challenging.

The available solutions for handling unobservable FDI attacks in power systems generally fall under two categories of matrix decomposition and statistical learning. The former models the problem as a type of matrix decomposition process \cite{7529082,7792204,8472173}. Devising matrix decomposition can facilitate the injected error estimation and the control signal retrieval through solving an optimization problem. However, the performance is usually dependent on the sparsity of attacks \cite{7792204}. Another approach with more generalization ability is reformulating the intrusion detection problem from the perspective of statistical learning \cite{8409487, 9207760, 6032057}. Using this technique, one can find the relationships between data samples w.r.t. different states such as normal and under attack. Statistical learning can also facilitate the classification of different types of data attacks, which can be informative for taking the right action when the system is under attack.

On the other hand, FDI attack detection and classification using data-driven approaches is followed by a number of challenges. To begin with, multi-class classification is the most common problem that is well studied in the literature \cite{4633363, 5444873}. Tackling with insufficient labeled samples is another obstacle, which has been recently addressed by means of semi-supervised learning \cite{8628992, 5444873, 4633363}. However, most of these studies consider a stationary environment, in which the feature space remains intact. This is while streaming phasor measurements in a power system resemble a non-stationary environment, where different factors such as system events and intrusions can alter the feature space. Such changes can be generally categorized into three types, namely concept drift \cite{6604410,5453372}, concept evolution \cite{5453372} and reoccurring classes \cite{7350165}. In non-stationary environments, the primary step is to detect any changes and anomalies in the data stream. The classification model can then be adapted to the detected changes using adaptive algorithms \cite{6604410,7332782}. Besides, most of the available solutions rely on external updates for adaptation, which is impractical in our case study. Each of the aforementioned problems, in both stationary and non-stationary environments, requires a different solution. Unfortunately, these problems are rarely addressed simultaneously for handling unobservable FDI attacks in power systems.

The aim of this work is to design a novel data-driven framework that aids in system state estimation, when the power system is under unobservable FDI attacks. In this process, we mainly rely on data mining and try to overcome all the aforementioned challenges for this approach. The proposed framework consists of three main modules, namely detection, classification, and control signal retrieval. The first module initially detects any unobservable FDI attacks and passes them to the attack classifier. The classification module then determines the type of these attacks. The last module retrieves the control signal and passes it to the control unit, using the patterns that have been captured by the other modules. Multiple challenges have been tackled in the design of the proposed framework including anomaly detection based on the historical behavior of the signal, extraction of attack patterns from the complex signals, scarce labeled data, concept drift, concept evolution, recurring classes, and fast signal retrieval.

The main contribution of this work is threefold: 
\begin{enumerate}
\item A new framework, un\underline{O}bservable th\underline{R}eat model\underline{I}n\underline{G} us\underline{I}ng least-squares fitti\underline{N}g (ORIGIN), is proposed for detecting unobservable FDI attacks in power systems. The novel idea behind ORIGIN is to model historical changes of PMU signal w.r.t. the deviation of the circular path it creates on a complex plane. For the normal signal, even if the radius of the signal changes, the path is still centered around the origin. Conversely, unobservable FDI attacks cause a deviation from the origin. To model the pattern of this deviation, we fit a least square circle to a sliding window on the streaming signal so that the resulted center can show the extent and direction of the deviation from the origin.
\item A novel ensemble learner, class\underline{I}fication using \underline{C}ross-c\underline{O}rrelatio\underline{N} (ICON), is proposed to facilitate unobservable FDI attack classification in non-stationary environments. ICON is well-adapted to the selected case study and can address all the mentioned challenges in this process. It performs semi-supervised multi-class classification, detects new classes of unobservable FDI attacks, performs self-update, and handles recurring classes. Unlike most of the available work, ICON addresses all these challenges simultaneously and does not rely on external updates.
\item The introduced signal retrieval method, that is used within ORIGIN, is also unique to this algorithm. It dynamically computes and removes the injected error from the control signal, in a sample-by-sample manner. Moreover, we evaluate the proposed methods by enabling a comparative study on real-world data, collected from the Central New York Power System (CNYPS).
\end{enumerate}

The remainder of this paper is organized as follows: Section \ref{sec:background} contains the background for this work. Section \ref{sec:design} explains the proposed framework. Experimental results are analyzed in Section \ref{sec:exp}. Finally, the paper is concluded in Section \ref{sec:conc}.

\section{Background}
\label{sec:background}
Here, we initially explain the considered problem and the selected case study. Then, the additional challenges of our case study, which resembles a non-stationary environment, and available methods to address them are elaborated.

\subsection{Problem Statement}
\begin{figure}[t]
\centering
\includegraphics[width=\columnwidth]{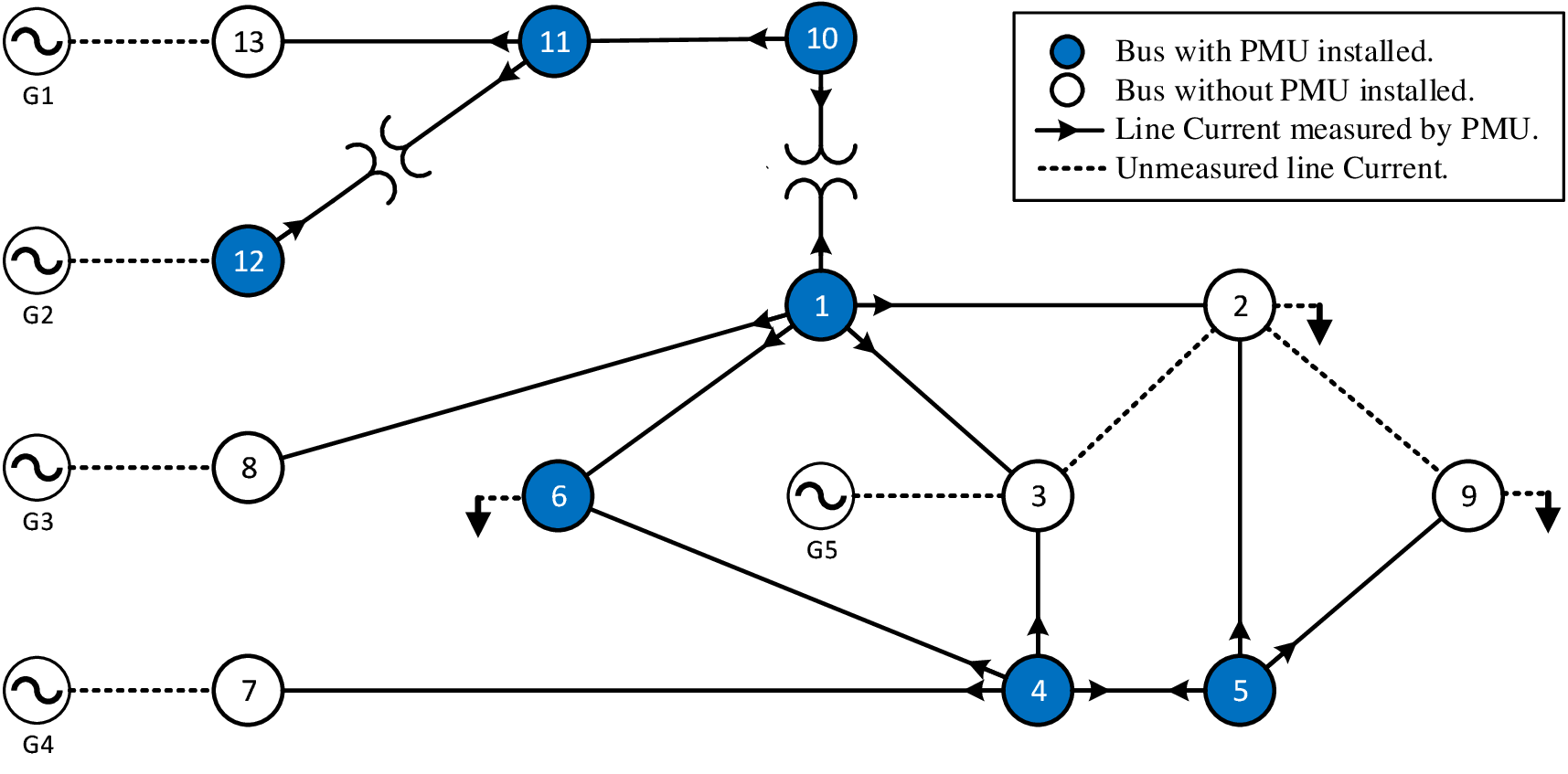}
\caption{Network topology of the Central New York Power System \cite{7529082}.}
\label{fig:network}
\end{figure}

Fig. \ref{fig:network} shows the structure of the CNYPS, on which the phasor measurement unit (PMU) data has been recorded and used in this work \cite{7529082}. Assuming that an intruder has partial knowledge of this system, data attacks can be planned by corrupting multiple PMU measurements. Failure to detect these attacks prevents the prompt retrieval of the control signal. This results in the malfunction of the control system.

Most of cyber data attacks are usually in the form of FDI attacks \cite{7313024,8472173}. Among various types of data attacks, unobservable FDI attacks are often known as the most challenging ones since they can bypass the bad data detectors (BDD) and manipulate the the control unit \cite{7529082}. In order to make the FDI attack unobservable to the BDD, the intruder must have at least partial knowledge of the system. Generally we assume that the attacker has control over a set of measurements and knows the network topology (at least partially). Appendix 1 formally explains how unobservable FDI attacks can bypass conventional, or residue-based, BDD. Unobservable attacks are particularly based on a corruption matrix $C$, which represents the additive error to the targeted bus voltages due to data attacks. Thus, an accurate IDS is necessary to know when the control signal misleads the control system. Once an attack is detected, a control signal retrieval process should be initiated to make the power system observable again.

Although a proper IDS and control signal retrieval scheme may usually guarantee the safe operation of the control unit, having more information about the strategy of the unobservable FDI attacks can help further secure the network. This means that detected unobservable attacks need to be classified into different categories in order to clarify the intention of an intruder. For instance, repetitious attacks with similar patterns  indicate an intruder has targeted a specific component of the network, and, thus, security protocols need to be enhanced for the respective component.

Hence, a threefold problem is addressed in this paper: a) unobservable FDI attack detection; b) classification of the detected attacks in non-stationary environments; and c) control signal retrieval. This is while the classification in a non-stationary environment, similar to that of CNYPS, has multiple additional challenges, that further complicate this process. These challenges are explained in the following subsection.

\subsection{Dynamic State Estimation}
A common trend often seen in the literature is to consider the steady state model in the design of monitoring applications at the control center of power systems \cite{book,4074022}. However, power systems rarely operate in a stationary or steady-state condition due to the stochastic variations in demand and generation. Therefore, this assumption becomes problematic since approaches using static state estimation cannot capture system dynamics in non-stationary environments \cite{8624411}. To address this issue, dynamic state estimation methods emerged and new monitoring tools were proposed in the wake of them \cite{8278264,8759957}. Dynamic state estimation enables capturing the history of dynamic events in power systems, which in turn facilitates the understanding and analysis of the behaviour, performance, and control decisions of the system.

\subsection{Model-based Approaches}
Dynamic state estimation using model-based methods requires the mathematical system model to be available. These methods usually apply estimation techniques to observe the state of a system and detect cyber-attacks based on analytical redundancies \cite{9409791}. Observer-based methods can be categorized into three groups, namely Kalman filtering \cite{8278264}, observer-based methods \cite{9318019, 9129765}, and sliding mode observation \cite{8752431}.

Kalman filtering techniques devise a state–space model and can employ an optimal recursive algorithm by means of previously estimated states and previous measurements. These techniques can accurately estimate the state vectors and are computationally inexpensive \cite{8278264}. Despite the effectiveness of Kalman filtering, they may not apply to some cases since they assume system and observation models equations are both linear. It is worth mentioning that the non-linear extension of Kalman filter is not often as optimal as the linear version. Moreover, it assumes that the state belief is Gaussian distributed.

Observer-based methods usually estimate the state of system by means of a deterministic state–space mathematical model \cite{9409791}. For instance, an methodology is proposed in \cite{9547709} for mitigating FDI attacks by considering unknown nonlinear constant power loads. This nonlinear and adaptive observer-based method also takes uncertainty, noise, and unobservable states into account. \cite{Saif2003} proposes an unknown input observer to detect FDI attacks by means of its residuals. A residual threshold is then considered to adaptively monitor the effect of disturbance. An unknown input interval observer-based approach is also designed in \cite{8957680} to defend against FDI attacks. The proposed method is a global detection algorithm and works based on the interval residuals. The main problem of observer-based methods is that the convergence and stability of closed-loop control system must be assured for all operating space. However, in control of non-linear systems these conditions are validated only locally or they are dependent of some hypothesis which reduce its effectiveness in real applications.

Sliding mode techniques are non-linear and devise a state-space model for estimating the state on a manifold w.r.t. estimated error and state trajectories. State estimation under this category comes with several benefits, including ease of use, disturbance rejection, and robustness \cite{article,7801908}. On the negative side, such methods may excite unmodeled dynamics and cause chattering. Therefore, instrumental damages and energy loss can take place in case of high-frequency switching around the sliding surface \cite{article}.

\begin{figure*}[t]
\centering
\includegraphics[trim={2.35cm 1.1cm 0.9cm 0.9cm},clip,width=\textwidth]{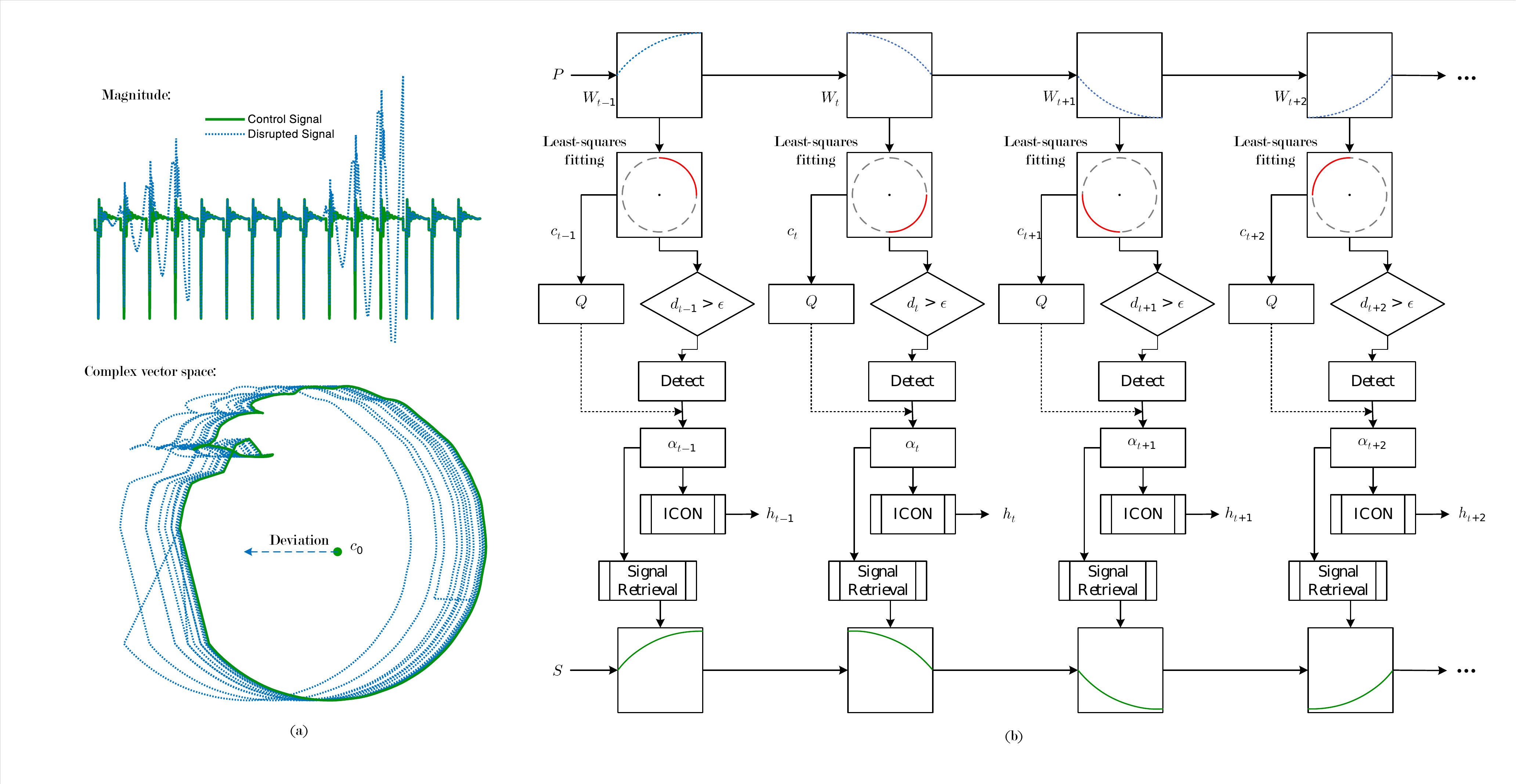}
\caption{Illustration of the designed hybrid framework. (a) shows how changes in the PMU measurements result in deviation of the estimated centers from the origin computed on the training data. The solid line denotes the clean control signal whereas the dotted line indicates the disrupted signal. (b) contains the block diagram of the ORIGIN framework.}
\label{fig:visio}
\end{figure*}

\subsection{Intelligent Data-driven Approaches}
As opposed to model-based techniques, intelligent data-driven methods are able to operate without any prior knowledge regarding the physics of the system, and the mathematical system model is not required \cite{9055170, 8519325, LI2020101705}. System state estimation using cyber-secure data-driven techniques mainly involves a classification task, which facilitates the detection of FDI attacks and the decision making process. Although intelligent methods for FDI detection and state estimation are have shown outstanding performance and flexibility in the literature, their main drawbacks are usually the assumption of a stationary environment or the high computational cost in non-stationary environments, where real-time model updating is required. However, this can be potentially addressed via an efficient design, and, thus, extensive research endeavours have been dedicated to address this issue \cite{7350165,5453372,8336956}.

\subsection{Challenges of Non-stationary Environments}
The classification procedure is often performed in a static manner, where a classification model is firstly constructed offline and, then, used in a dynamic environment. This setting works well for stationary environments, where the data flow resembles a feature space similar to that of the constructed model. However, in the intrusion detection problem, an intrusion may bring about unexpected anomalies in the upcoming data, whether gradually or abruptly, which alter the mapping between the input samples and class labels. This will consequently deteriorates the performance of the trained model, as it cannot interpret the new data generating process. This phenomenon is known as concept drift. Note that the standard i.i.d. assumptions do not hold anymore in non-stationary data steams.

Another issue of concern is the presence of new classes of attacks in the network. The presence of a new class in feature space is also referred to as concept evolution \cite{7350165,5453372}. Failure to detect new classes of attacks results in misclassifying attack patterns and generation of incorrect information. Various research efforts have been dedicated to the real-time classification in non-stationary environments \cite{6604410,5453372}. The problem has been adequately addressed in these works; however, they are not designed to handle reoccurring classes. This means that if a class of attack is absent for a while, the dynamic classification model will forget that class, and it will be detected as a new class when it reoccurs. This problem may be eliminated by resorting to an active learning setting \cite{7762928}, in which a request for an external update is sent to an operator whenever an algorithm cannot decide confidently about a sample. Nevertheless, dependency to human experts for the intrusion detection is a time consuming and expensive task.

A more realistic approach is then presented in \cite{7350165}, called \underline{Cla}ss-based ense\underline{m}ble (CLAM), which relies on periodical updates rather than waiting for an operator response. It uses an ensemble of one-class classifiers that models clusters of data by means of their centroid and radius. CLAM detects new classes using silhouette coefficients and updates the classification model. \underline{S}tream \underline{C}lassifier \underline{A}nd \underline{N}ovel and \underline{R}ecurring class detector (SCANR) \cite{6137334} is another algorithm proposed for addressing concept evolution and recurring class, which uses the same approach as CLAM for modelling the data clusters. However, it makes use of a primary ensemble for classification and an auxiliary ensemble for detecting recurring classes.

Although CLAM and SCANR address most of the aforementioned challenges, there are still a few factors that limit their use for the application of intrusion detection. To begin with, dependency on human-generated periodical updates is not preferable for this application since an attack should be isolated as soon as it is detected. Secondly, relying only on cluster-based modeling may not be the best strategy for cases such as PMU data, in which samples are presented using complex numbers. These issues motivated us to propose ICON especially for unobservable FDI attack classification, which can address all the aforementioned problems independently. Note that many classifiers that use computationally expensive approaches, e.g., deep learning, are not compatible with this case study, as they cannot be rapidly retrained or updated.

\section{Design of the Intrusion Detection and Classification System}
\label{sec:design}
Here, we formally explain the the proposed framework, which is also detailed in Algorithm \ref{alg:alg1} and illustrated in Fig. \ref{fig:visio}. Supposing that data is being captured through a set of PMUs $p = \{P_1,P_2,...,P_{\varsigma}\}$, each PMU $P$ contains $\vartheta$ channels resulting in a data matrix $M\in \mathbb{C}^{n\times \vartheta}$, where $n$ is the number of samples.

\subsection{Dynamic Intrusion Detection using ORIGIN}
Assuming that $M_j$ represents the $j$-th column of $M$, $M_j = \{z_i\}^{n}_{i=1}$, where $z_i\in\mathbb{C}$. For a set of points in a sliding window $W = \{z_1,z_2,...,z_{\omega}\}$, where $\omega<n$, a circle can be algebraically fitted to $W$. Considering the algebraic representation of a circle $F(\mathbf{x})=0$ for a set of coordinate vectors $\mathbf{x}\in W$, where $\mathbf{x}$ consists of real and imaginary parts of samples in $W$, an algebraic fit is attained by determining the parameters of this equation w.r.t. least-squares. By doing so, the estimated circular path via the least-square fitting reflects the historical changes in the dynamic system, where older samples locate at the end of this window, and vice versa. 

\begin{assumption}\label{as:as1}
Given that a center $c_0=[x_0,y_0]$ can be estimated by fitting a circle to training samples $P_0$ with $n_0$ samples (see Fig. \ref{fig:visio}.a), it is assumed that when the system is not under attack or injected error $\Vert e\Vert=0$ then $\Vert c_0 - c_t\Vert \leq \delta$, where $c_t$ is the estimated centers w.r.t. the sliding window $W_t$ at time $t$, and $\delta$ is a threshold indicating the degree of center deviation when $\Vert e\Vert=0$.
\end{assumption}

Intrusion detection can then be performed based on the following proposition, which is proved in the Appendix 2 of this paper.
\begin{prop}\label{th:th1}
The presence of an unobservable FDI attack at time $t$, when $\Vert e_t\Vert > \Vert\delta\Vert$, is sufficient to deviate $c_t$ in an extent that exceeds the threshold $\delta$.
\end{prop}

\begin{algorithm}[t]
\caption{ORIGIN}
\label{alg:alg1}
\KwIn{\\PMU data $P$, threshold $\delta$.}
\KwOut{\\Intrusion detection result $\mathcal{F}$, attack classification result $h$, retrieved control signal $S$.}
\SetKwInput{Initialization}{Initialization}
\SetKwInput{Definition}{Definition}
\SetKwInput{Testing}{Testing}
\SetKwInput{Training}{Training}
\Definition{\\$X_0$ is the training set, $\leftarrow$ inserts into a FIFO queue.}
\Initialization{}
Initialize FIFO queue $Q=\varnothing$.\\
Set intrusion detection flag $\mathcal{F}=False$.\\
Set counter $j=1$.\\
\Training{}
Compute $c_0$ on $P_0$ using Equations (\ref{eq:eq1})-(\ref{eq:eq5}).\\
\Testing{}
\While{$P$}{
Update the sliding window $W_t = P(r)$, $j\leq r< j+\omega$.\\
Compute $\hat{c}_t$ on $W_t$ using Equations (\ref{eq:eq1})-(\ref{eq:eq5}).\\
$j = j+1$. \\
$d_t=\Vert c_0 - \hat{c}_t \Vert$.\\
Update FIFO queue $Q\leftarrow d_t$.\\
\If{$d_t > \delta \wedge \operatorname{Card}(Q)=\tau$}{
Set intrusion detection flag on $\mathcal{F}=True$.\\
Create an attack sample $\alpha_t = Q$.\\
Call the attack classifier $h_t=\operatorname{ICON}(\alpha_t)$.\\
$\forall \alpha_t$ compute $M_{rj}$ using Equation (\ref{eq:eq10}) to retrieve the control signal S.
}
}
\end{algorithm}

Thus, we first create a model using the original center $c_0$ in the training set (Line 4 in Algorithm \ref{alg:alg1}). Then, at each time-step $t$, the sliding window $W_t$ moves on the incoming data (see Fig. \ref{fig:visio}.b), and $\hat{c}_t$ is computed on the current window (Line 6 and 7 in Algorithm \ref{alg:alg1}). After computing the deviation $d_t$ between these two coordinates, it is added to a First-In-First-Out (FIFO) queue $Q$ (Lines 9 and 10 in algorithm \ref{alg:alg1}). Once $Q$ reaches its capacity, an intrusion is being detected any time $d_t>\delta$ (Lines 10 and 11 in Algorithm \ref{alg:alg1}). Thereafter, $Q$ is exported as an attack sample and sent to the attack classifier module, which will be explained shortly, and the control signal is retrieved afterwards (Lines 13-15 in Algorithm \ref{alg:alg1}).

\subsection{Online Modeling of Attack Patterns in ORIGIN}
In order to classify the detected attack, the model needs to capture of the changes each type of unobservable FDI attack makes in the system. For instance, the injected errors can increase or decrease the signal magnitude, with various ratios. Thus, variations of unobservable FDI attack create unique patterns in the PMU signal based on these variables. Henceforth, we refer to these historical patterns as attack patterns.

Once an attack is detected, the attack pattern should be modeled properly in order to create a set of useful samples for attack classification. To do so, a FIFO queue $Q$ is considered with a constant size $\tau$ (see Fig. \ref{fig:visio}.b). At each time-step $t$, a scalar $d_t=\Vert c_0 - \hat{c}_t\Vert$ is stored in $Q$. When $\operatorname{Card}(Q)=\tau$, where $\operatorname{Card}(\cdot)$ gives the cardinality of a set, an attack pattern $\alpha$ is given by extracting the sequence in $Q$ (Line 14 in Algorithm \ref{alg:alg1}). This pattern shows the deviation of $c_t$ over time (see Fig. \ref{fig:visio}.a). These generated samples, or attack patterns, are then fed to the attack classification module, for which variant attack patterns are indicators of different classes.

\subsection{Real-Time Attack Classification via ICON}
Initially, an offline training phase is performed to create a base model for attack classification (Line 3-8 in Algorithm \ref{alg:alg2}). Afterwards, the model keeps getting updated automatically during the attack classification, if needed. 

Firstly, an empty ensemble $E=\varnothing$ is initialized (Line 1 in Algorithm \ref{alg:alg2}). For a set of training samples $X_0 = \{\alpha_1,\alpha_2,...,\alpha_{n_0}\}$, $\alpha_1$ is selected to create the first class $\varepsilon_1$ (Line 2 in Algorithm \ref{alg:alg2}). For $\alpha_r$, $2<r\leq n_0$, the cross-correlation of $\alpha_r$ and $\varepsilon_j\in E$, where $1\leq j\leq\nu_t$ and $\nu_t$ is the number of classes until time $t$, is computed (Line 4 in Algorithm \ref{alg:alg2}). We consider the symmetry of this cross-correlation as the similarity measure:
\begin{equation}
\label{eq:eq6}
\Xi_{rj}(m) = \begin{cases}
\sum\limits_{i=0}^{N-m-1}{\alpha_r(i+m)\varepsilon_j^*(i)},~~ m\geq 0.\\
\Xi^*_{jr}(-m), \qquad\qquad\qquad~ m<0.
\end{cases},
\end{equation}
where the asterisk indicates complex conjugation, $(\cdot)$ is used for indexing on a sequence, and $N$ is the length of the pattern in $\alpha_r$ and $\varepsilon_j$. Then, the output vector $\rho$ is resulted from $\Xi\in \mathbb{R}^{1\times N^2}$, where $\rho_i = \Xi_{rj}(m -N)$ and $1\leq m\leq 2N-1$, $m\in\mathbb{N}$ (Line 5 in Algorithm \ref{alg:alg2}). The similarity measure $\upsilon$ is then calculated (Line 6 in Algorithm \ref{alg:alg2}) as follows:
\begin{equation}\label{eq:eq7}
\upsilon =\bigg\vert\sum\limits_{i=1}^\kappa{\rho_i}-\sum\limits_{i=\kappa+1}^{2N-1}{\rho_i}\bigg\vert,~~\kappa=\bigg\lceil{\frac{\operatorname{Card}(\rho)}{2}}\bigg\rceil.
\end{equation}

A parameter $\gamma\in\mathbb{R}^+$ is then considered to adjust the sensitivity of the classifier. By this mean, $\forall\alpha_r\in X_0:$
\begin{equation}
\label{eq:eq8}
\begin{cases}
E_0 \cup \varepsilon_{new} : \varepsilon_{new} = \alpha_r, ~~ \upsilon>\gamma.\\
\varepsilon_i = \varepsilon_i \cup \alpha_r : \varepsilon_i \in E_0, ~~~ \upsilon \leq \gamma \wedge Card (\varepsilon_i) < \lambda.\\
\varepsilon_i = \{\varepsilon_i \bigtriangleup \alpha_{out}\}  \multirow{2}{*}{\qquad,~\quad$\upsilon \leq \gamma \wedge Card (\varepsilon_i) \geq \lambda.$} \\ \qquad\quad \cup~ \alpha_r:\varepsilon_i \in E_0
\end{cases},
\end{equation}
where $\lambda$ is the memory size of each class model $\varepsilon$, $\varepsilon_{new}$ denotes to a new model for a new class, and $\alpha_{out}\in \varepsilon_i$ is the oldest pattern in the class model (Line 7 in Algorithm \ref{alg:alg2}).

Once the ensemble is trained, it can operate by receiving new samples. At each time-step $t$, $E_t-1$ is used to start the classification (Line 10 in Algorithm \ref{alg:alg2}). Then, similar to Equation (\ref{eq:eq8}), if a new class is detected, a new-class label is assigned to the attack sample and a new model is assigned to it before updating $E_t$ (Lines 11-13 in Algorithm \ref{alg:alg2}). Otherwise, the detected attack is classified under its most similar class of attack and the associated class-memory will be updated (Lines 14-21 in Algorithm \ref{alg:alg2}). Finally, the classification result will be returned to ORIGIN.

\begin{algorithm}[t]
\caption{$\operatorname{ICON}$}
\label{alg:alg2}
\KwIn{\\Attack sample $\alpha_t$, sensitivity parameter $\gamma$.}
\KwOut{\\Attack classification result $h_t$.}
\SetKwInput{Initialization}{Initialization}
\SetKwInput{Testing}{Testing}
\SetKwInput{Training}{Training}
\SetKwInput{Definition}{Definition}
\Definition{\\$X_0$ is the training set, $\leftarrow$ inserts into a FIFO queue, $\operatorname{Label}(\cdot)$ returns the label.}
\Initialization{}
Initialize an empty ensemble $E=\varnothing$.\\
Initialize first class-memory $\varepsilon_1=\alpha_1$.\\
\Training{}
\For{$\forall \alpha_r$ , $2\leq r\leq n_0$}{
Compute $\Xi_{rj}(m)$ using Equation (\ref{eq:eq6}).\\
$\rho_r = \Xi_{rj}(m-N)$, $1\leq m\leq 2N-1 \wedge m\in \mathbb{N}$.\\
Calculate similarity measure $\upsilon$ using Equation (\ref{eq:eq7}).\\
Update $E_0$ using Equation (\ref{eq:eq8}).\\
}

\Testing{}
\While{$X_t\neq \varnothing$}{
$E_t = E_{t-1}$.\\

\uIf{$\upsilon>\gamma$}{
$h_t=\operatorname{Label}(\varepsilon_{new})$.\\
$E_t \cup \varepsilon_{new} : \varepsilon_{new} = \alpha_i$.}
\Else{
$h_t$ = $\operatorname{Label}(\varepsilon_i)$.\\
\uIf{$Card (\varepsilon_i) < \lambda$}{
$\varepsilon_i = \varepsilon_i \cup \alpha_i : \varepsilon_i \in E_t$.}
\Else{$\varepsilon_i = \{\varepsilon_i \bigtriangleup \alpha_{out}\}\cup \alpha_i: \varepsilon_i \in E_t$.}}
\Return{$h_t$}
}
\end{algorithm}

\subsection{Control Signal Retrieval}
Once the attack sample is classified, the original phasor measurement can be estimated as in the following:
\begin{equation}\label{eq:eq9}
\begin{split}
\forall \alpha_r: M_{rj}& = x_r - \left(\frac{1}{\tau}\sum\limits_{\iota=1}^{\tau}d_\iota\right)\cos\left(\tan^{-1}\left(\frac{x_t}{y_t}\right)\right) +\\& i\left(y_r -\left(\frac{1}{\tau}\sum\limits_{\iota=1}^{\tau}d_\iota\right)\sin\left(\tan^{-1}\left(\frac{x_t}{y_t}\right)\right)\right)
\end{split},
\end{equation}
which can be further simplified to the following form:
\begin{equation}\label{eq:eq10}
M_{rj} = x_r + iy_r - \frac{1}{\tau\sqrt{1+\theta^2}}\sum\limits_{\iota=1}^{\tau}d_\iota (1+i\theta^2),
\end{equation}
where $\theta=x_t/y_t$. Therefore, the control signal is retrieved in a sample by sample manner and without delay. This is mainly due to the usage of overlapping windows. In other words,  when a new sample arrives, $W_t$ moves forward for one sample. However, in Fig. \ref{fig:visio}.b windows are merely shown for non-overlapping parts of the signal for the sake of simplicity.

While the proposed framework is compatible with all power systems, there are a few points to be considered for optimal adaptation to different systems. Firstly, the ORIGIN method requires the input data to be in the form of sinusoidal waveforms (e.g., PMU signal). Secondly, for each selected PMU, the $\delta$ and $\gamma$ parameters should be adjusted for ORIGIN and ICON, respectively.

\section{Experimental Results}
\label{sec:exp}
We first describe the captured PMU data and the designed scenarios. Then the experimental setting is explained prior to the analysis of results.

\subsection{PMU Data and Scenarios}
\begin{figure}[t]
\centering
\includegraphics[trim={1.3cm 0cm 1.52cm 0.5cm},clip,width=\columnwidth]{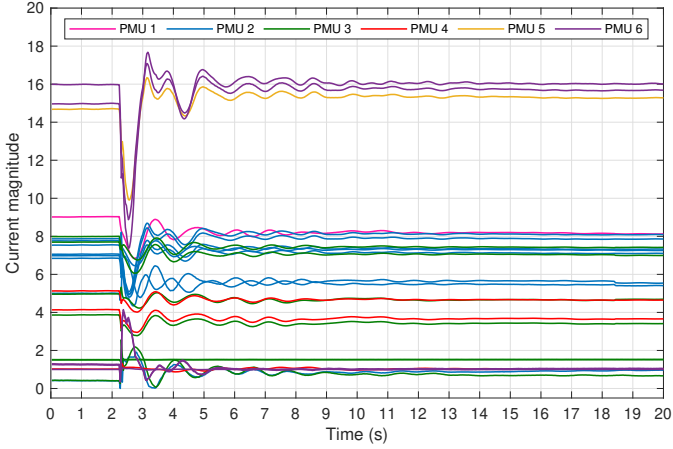}
\caption{Visualization of current magnitudes in the PMU data.}
\label{fig:current}
\end{figure}

In this work, we use real-world PMU data that were recorded through the CNYPS. The data is captured within a 20-$s$ period with a 30 $Hz$ sampling rate resulting in 600 samples. Six PMUs are used for this purpose, which contain 37 channels in total. Here, four scenarios are designed in which the intruder select different numbers of PMUs and channels are selected to launch unobservable FDI attacks, as listed in Table \ref{tab:scenario}. Table \ref{tab:scenario} also shows the phasor measurements for both voltage and current in each scenario. In this data, an event  (i.e., a slip in a generator) takes place around two seconds, which is observable in Fig. \ref{fig:current}.

\begin{figure}[t]
\centering
\includegraphics[trim={1.1cm 1cm 1cm 1.2cm},clip,width=\columnwidth]{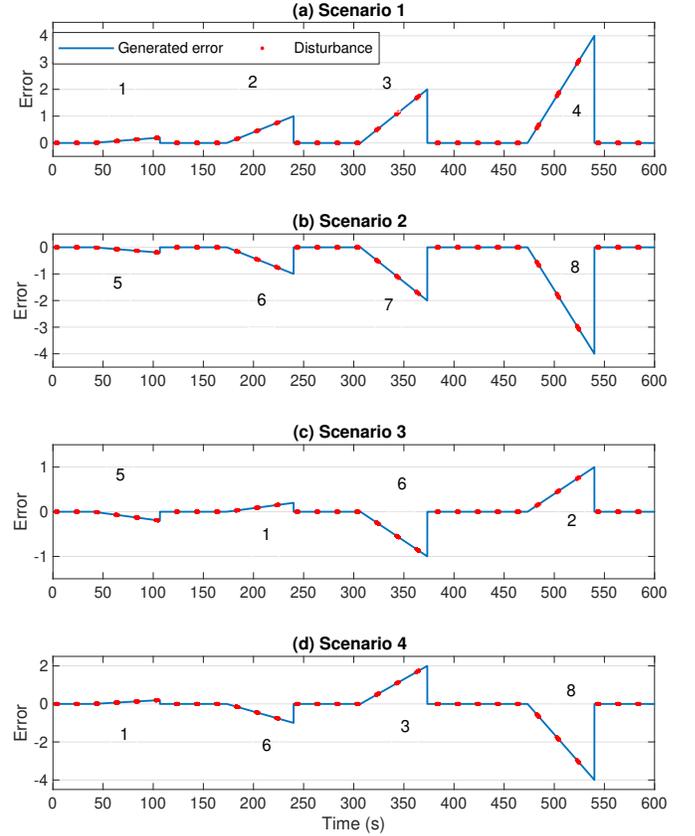}
\caption{Generation of unobservable attacks in four different strategies. Each strategy consists of four different classes of attack that are shown with numbers around the generated errors.}
\label{fig:Attack}
\end{figure}

In order to simulate longer scenarios, the described data has been repeated for thirty cycles. The unobservable attacks are then injected to the data numerically through the corrpution matrix $C$ (see Appendix 1). There are different approaches to achieve $C$. For instance, \cite{7529082} formulates $C$ randomly selecting column supports and fills non-zero values with Gaussian noise. \cite{8587555} uses a diagonal full-rank matrix in which indices corresponding to the targeted buses will be replaced with the exponent of the product of the index and a scalar. times the column index of the targeted bus. Similarly, our work uses the replacement of indices of $C$ that correspond to the targeted buses w.r.t. the network topology. Four different strategies are considered for attempting an unobservable attack to the power system, as shown in Fig. \ref{fig:Attack}. The first strategy contains only additive errors (see Fig. \ref{fig:Attack}.a), while the second one consists of deductive errors (see Fig. \ref{fig:Attack}.b). The last two strategies, on the other hand, blend both additive and deductive classes of attack listed in Table \ref{tab:attack}. As illustrated in Fig. \ref{fig:Attack}.c, the third strategy increases the error magnitude equally for additive and deductive errors, whereas in the last one the error magnitude increases for each class of attack regardless of the error type (see Fig. \ref{fig:Attack}.d).

\begin{table}[t]
\centering
\caption{Characteristics of the designed scenarios.}
\tabcolsep 15.5pt
\begin{tabular}{lcccc}
\hline
\multirow{2}{*}{Scenarios} & \multirow{2}{*}{\#. PMUs} & \multicolumn{3}{c}{\#. Channels}\\
\cline{3-5}
& & Total & $V$ & $I$\\
\hline
Scenario 1 & 3 & 16 & 6 & 10\\
Scenario 2 & 3 & 21 & 5 & 16\\
Scenario 3 & 4 & 23 & 7 & 17\\
Scenario 4 & 4 & 27 & 7 & 20\\
\hline
\end{tabular}
\label{tab:scenario}
\end{table}

\begin{table}[t]
\centering
\caption{Magnitude of additive or deductive errors for each class of attack that should be injected by an intruder per $1/30$ $s$.}
\label{tab:attack}
\tabcolsep 7.5pt
\begin{tabular}{lcccccccc}
\hline
Attack class & 1 & 2 & 3 & 4 & 5 & 6 & 7 & 8\\
\hline
Error ($\times 10^{-3}$) & 0.1 & 0.5 & 1 & 5 & 0.1 & 0.5 & 1 & 5 \\
Additive &  \checkmark & \checkmark & \checkmark & \checkmark & & & &\\
Deductive & & & & & \checkmark & \checkmark & \checkmark & \checkmark\\
\hline
\end{tabular}
\end{table}

\begin{figure}[t]
\centering
\includegraphics[trim={0.8cm 0.3cm 1cm 0.4cm},clip,width=\columnwidth]{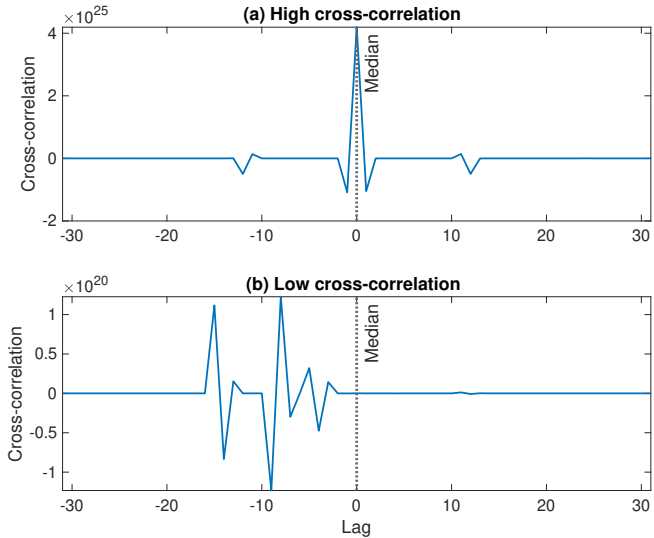}
\caption{Visualization of cross-correlation sequence between two random pairs of samples in the training set. (a) shows a symmetric pattern resulted by high cross-correlation, while (b) illustrates an asymmetric pattern resulted by low cross-correlation.}
\label{fig:corr}
\end{figure}

\begin{table}[!t]
\centering
\caption{Attack detection results over four different scenarios. Accuracy measurements are reported in percentage.}
\label{tab:detection}
\tabcolsep 5.5pt
\begin{tabular}{clccccc}
\hline
 \multicolumn{2}{c}{\multirow{2}{*}{Experiments}} & \multirow{2}{*}{Accuracy} & \multicolumn{2}{c}{Safe} & \multicolumn{2}{c}{Intrusion}\\
\cline{4-7}
& & &Precision & Recall & Precision & Recall\\
\hline
\multirow{4}{*}{\rotatebox[origin=c]{90}{ORIGIN}}& Scenario 1 & 97.4 & 99.3 & 96 & 95.2 & 99.1\\
& Scenario 2 & 97.4 & 99.4 & 96 & 95.2 & 99.2\\
& Scenario 3 & 95.8 & 96.7 & 95.7 & 94.7 & 95.9 \\
& Scenario 4 & 97.3 & 99.1 & 96 & 95.2 & 99 \\
\hline
\multirow{4}{*}{\rotatebox[origin=c]{90}{DM}}& Scenario 1 & 92.1 & 93.8 & 90.7 & 90 & 93.7\\
& Scenario 2 & 87.2 & 88.7 & 86.1 & 85.3 & 88.9\\
& Scenario 3 & 84.1 & 85 & 84.9 & 83.2 & 84.2\\
& Scenario 4 & 79.5 & 81.1 & 78.9 & 77.9 & 80.9 \\
\hline
\end{tabular}
\end{table}

\subsection{Experimental Setting}
All experiments are conducted within a 10-fold stratified cross-validation procedure, which consists of an inner and an outer loop. The outer loop keeps one fold for testing and the rest for the training, while the inner loop performs another cross-validation procedure on the training folds to find the optimal parameters.

CLAM, SCANR, and Fast-COMPOSE \cite{8336956} are provided with periodical updates, as it is required by them, since their performance would decrease drastically without updates.

\subsection{Results Analysis}
The results are reported and analyzed separately for each module.

\begin{table*}[t]
\centering
\caption{Averaged class specific performance measures ($\%$) for attack classification w.r.t. each scenario obtained through a 10-fold cross-validation.}
\label{tab:isolation}
\tabcolsep 4pt
\begin{tabular}{|l|c|ccc|ccc|ccc|ccc|}
\hline
\multirow{2}{*}{Scenarios} & \multirow{2}{*}{Attacks} & \multicolumn{3}{c|}{ICON} & \multicolumn{3}{c|}{CLAM} & \multicolumn{3}{c|}{SCANR} & \multicolumn{3}{c|}{Fast-COMPOSE}\\
\cline{3-14}
& & Accuracy & Precision & Recall & Accuracy &Precision & Recall & Accuracy &Precision & Recall & Accuracy &Precision & Recall\\
\hline
\multirow{5}{*}{Scenario 1} & Class 0 & 97.67& 99.29 & 96.11 &96.02& 97.64  & 94.46 &92.87&94.44& 91.36 & 88.2265	&	89.718	&	86.792
\\
& Class 1 & 96.57 & 94.32 & 98.93 &94.92& 92.67 & 97.29 &91.76&89.37& 94.29&	87.172	&	84.9015	&	89.5755
\\
& Class 2 & 96.95 & 94.91 & 99.08 & 95.16 & 93.27 & 97.13 & 91.96 &90.07&93.94&	87.362	&	85.5665	&	89.243
\\
& Class 3 & 97.13 & 95.18 & 99.18 &95.50& 93.55 & 97.55 &92.23&90.25&94.31&	87.6185	&	85.7375	&	89.5945
\\
& Class 4 & 97.95& 96.41 & 99.54 &96.28& 94.76& 97.85 &93.06&91.52&94.66&	88.407	&	86.944	&	89.927
\\

\hline
\multirow{5}{*}{Scenario 2} & Class 0 &97.77& 99.37 & 96.23 &96.14& 97.75 & 94.59 &93.67&95.81&91.64&	88.9865	&	91.0195	&	87.058
\\
& Class 5& 96.61& 94.38 & 98.96 &94.98& 92.75&97.32&92.09&89.91&94.38&	87.4855	&	85.4145	&	89.661
\\
& Class 6& 97.02 & 95.01 & 99.12 & 95.37 & 93.36& 97.48& 92.45&90.45&94.55&	87.8275	&	85.9275	&	89.8225
\\
& Class 7&97.16 & 95.20 & 99.21 & 95.52 & 93.58 & 97.56 &92.71&90.63&94.89&	88.0745	&	86.0985	&	90.1455
\\
& Class 8& 97.93& 96.34 & 99.58 &96.30& 94.71 & 97.95 &91.90&91.79&92.03&	87.305	&	87.2005	&	87.4285
\\

\hline
\multirow{5}{*}{Scenario 3} & Class 0&96.19& 96.69 & 95.71 &94.99& 95.48 & 94.51 &90.26&90.75&89.78&	85.747	&	86.2125	&	85.291
\\
& Class 5& 95.15& 94.41 & 95.91 &93.92& 93.21 & 94.66 &89.12&88.34&89.93&	84.664	&	83.923	&	85.4335
\\
& Class 1& 95.10 & 94.38 & 95.84 & 93.87 & 93.14 & 94.63 & 89.15 &88.41&89.91&	84.6925	&	83.9895	&	85.4145
\\
& Class 6& 95.58& 95.10 & 96.08 &94.36& 93.86 & 94.87 &91.66&91.16&92.17&	87.077	&	86.602	&	87.5615
\\
& Class 2& 95.57& 95.01 & 96.14 &94.35& 93.80 & 94.91 &92.20&92.23&92.17&	87.59	&	87.6185	&	87.5615
\\

\hline
\multirow{5}{*}{Scenario 4} & Class 0&97.61& 99.19 & 96.09 &95.88& 97.43 & 94.39 &92.67&94.21&91.18&	88.0365	&	89.4995	&	86.621
\\
& Class 1&96.55 & 94.38 & 98.84 &94.80& 92.62 & 97.09 &91.59&89.43&93.87&	87.0105	&	84.9585	&	89.1765
\\
& Class 6& 96.93 & 95.10 & 98.98 & 95.26 &93.37 &97.23& 92.05 &90.17& 94.01&	87.4475	&	85.6615	&	89.3095
\\
& Class 3& 97.09& 95.21 & 99.05 &95.34&93.46& 97.31&92.17&90.26&94.17&	87.5615	&	85.747	&	89.4615
\\
& Class 8& 97.87& 96.42 & 99.38 &96.13&94.68&97.63&92.96&91.48&94.49&	88.312	&	86.906	&	89.7655
\\
\hline
\end{tabular}
\end{table*}

\subsubsection{Intrusion Detection}
Table \ref{tab:detection} shows the result of attack detection of ORIGIN and the matrix decomposition approach (DM) over four different scenarios. Overall accuracy, precision, and recall (i.e., sensitivity) are used as performance measures in this work, as listed in Table \ref{tab:detection}.

The reported recall values in Table \ref{tab:detection} indicate the high sensitivity of ORIGIN to the presence of unobservable attacks. The attained precision values, on the other hand, display high true positive prediction rates, which imply a very low percentage of false alarms for the intrusion detection module. ORIGIN outperforms DM w.r.t. any of the considered performance measures.

Compared to ORIGIN, the main disadvantage of the traditional detectors that are based on the MD process is that their performance is highly bound to the sparsity of the attacks. In other words, as the number of channels under attack increases, the success rate of these detectors is more likely to deteriorate. For instance, \cite{7529082} uses a DM-based detection approach. While this method has shown somewhat acceptable robustness against moderate sparsity on the NYPS case study, its performance drops for high degrees of sparsity. Table \ref{tab:detection} clearly shows this issue for scenarios one to four, in which  47, 60, 65, and 82 percent of the channels are under attack, respectively. In contrast, ORIGIN is completely robust against the sparsity of attacks and does not show any sensitivity in this sense.

\begin{figure}[t]
\centering
\includegraphics[trim={1.5cm 0.7cm 1.4cm 0.45cm},clip,width=\columnwidth]{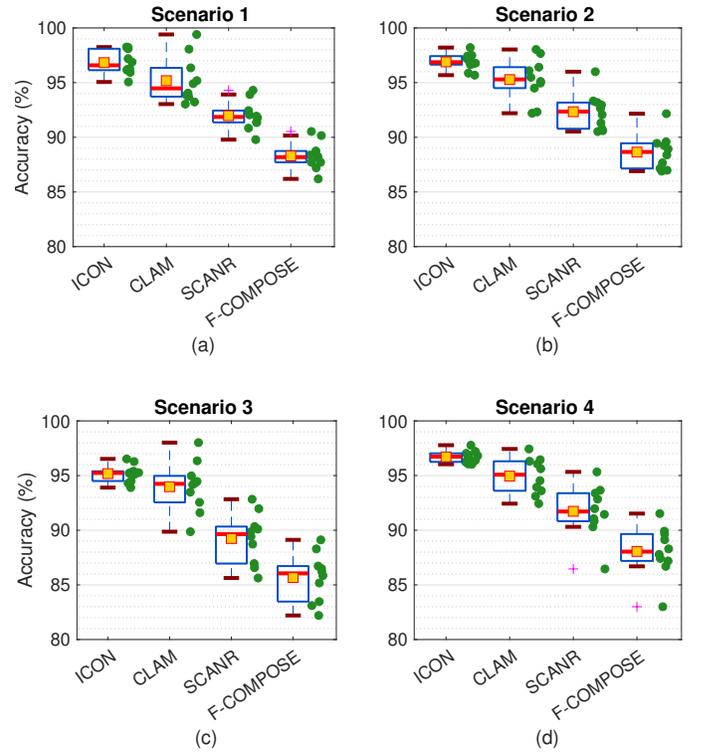}
\caption{Boxplot of the achieved accuracies for attack classification within each scenario. Solid dots shows the attained accuracies in each iteration of cross-validation, while the solid square denotes the mean of these accuracies.}
\label{fig:box}
\end{figure}

\begin{figure}[t]
\centering
\includegraphics[trim={0.9cm 0.5cm 1.15cm 0.6cm},clip,width=\columnwidth]{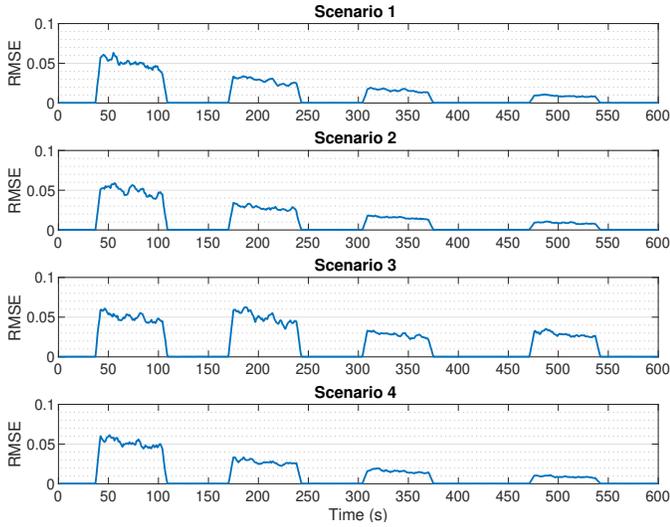}
\caption{RMSE of signal retrieval over time for four different scenarios.}
\label{fig:signal}
\end{figure}

\subsubsection{Attack Classification}
Fig. \ref{fig:corr} clarifies the intention behind using the symmetry of cross-correlation sequence as a similarity measures for comparing generated attack samples. It can be seen that the more a pair of samples are similar (see Fig. \ref{fig:corr}.a) the more symmetry is attained, and vice versa (see Fig. \ref{fig:corr}.b).

Fig. \ref{fig:box} shows the result of the cross-validation for attack classification in each scenario. Considering the distribution of obtained accuracies by each algorithm, it can be inferred that ICON is the most stable method among the others. Moreover, the average accuracy of ICON, denoted by a solid square in Fig. \ref{fig:box}, is always higher than CLAM, SCANR, and Fast-COMPOSE. This is while CLAM is ranked second in terms of accuracy and stability. However, the stability difference between CLAM and SCANR is negligible, and Fast-COMPOSE is ranked as fourth.

A more detailed report about the achieved performance measures for the conducted experiments can be found in Table \ref{tab:isolation}. Looking into the resulted accuracies, ICON shows superior performance for the task of attack classification, and it is ranked as first. Moreover, precision and recall values in Table \ref{tab:isolation} indicate higher sensitivity of ICON to all classes of unobservable FDI attacks in the designed scenarios. This is while, CLAM, SCANR, and Fast-COMPOSE are ranked from second to fourth, respectively.

\begin{figure}[t]
    \centering
    \includegraphics[width=\columnwidth]{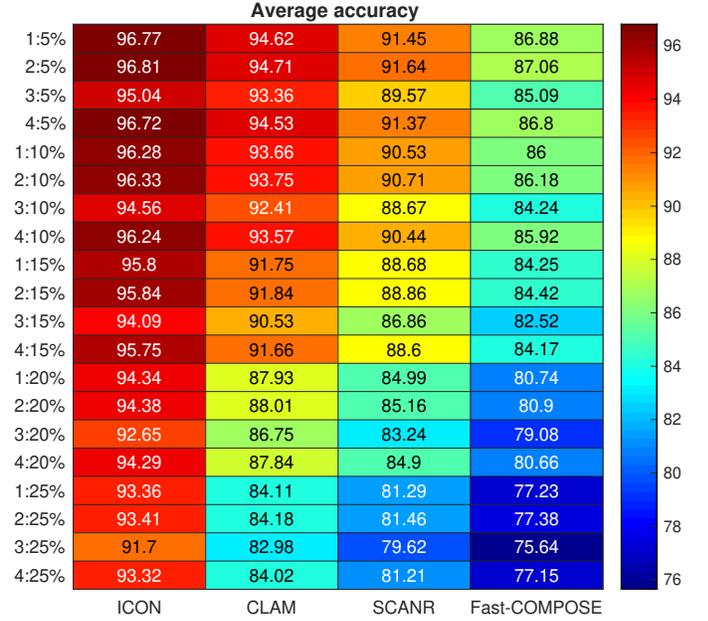}
    \caption{Average accuracy w.r.t. noise degree $\sigma$. Labels on the Y-axis show the scenario number versus $\sigma$.}
    \label{fig:heat_acc}
\end{figure}

\begin{figure}[t]
    \centering
    \includegraphics[width=\columnwidth]{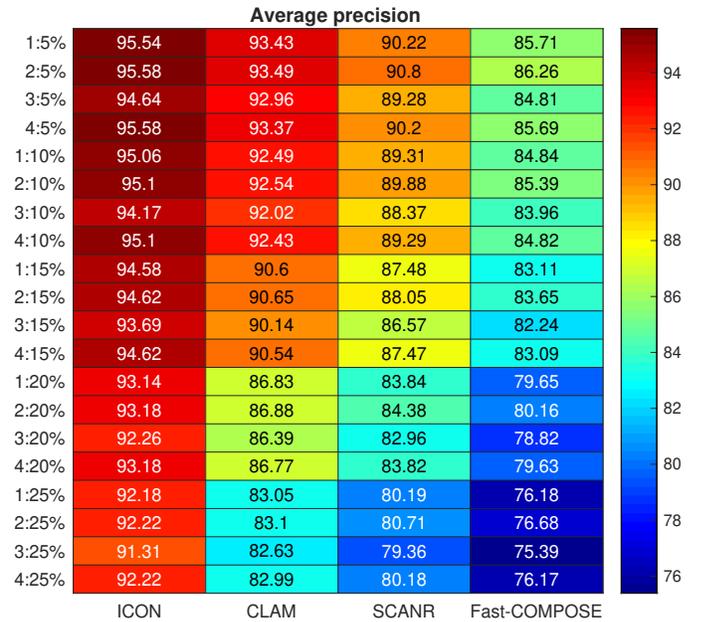}
    \caption{Average precision w.r.t. noise degree $\sigma$. Labels on the Y-axis show the scenario number versus $\sigma$.}
    \label{fig:heat_pre}
\end{figure}

\begin{figure}[t]
    \centering
    \includegraphics[width=\columnwidth]{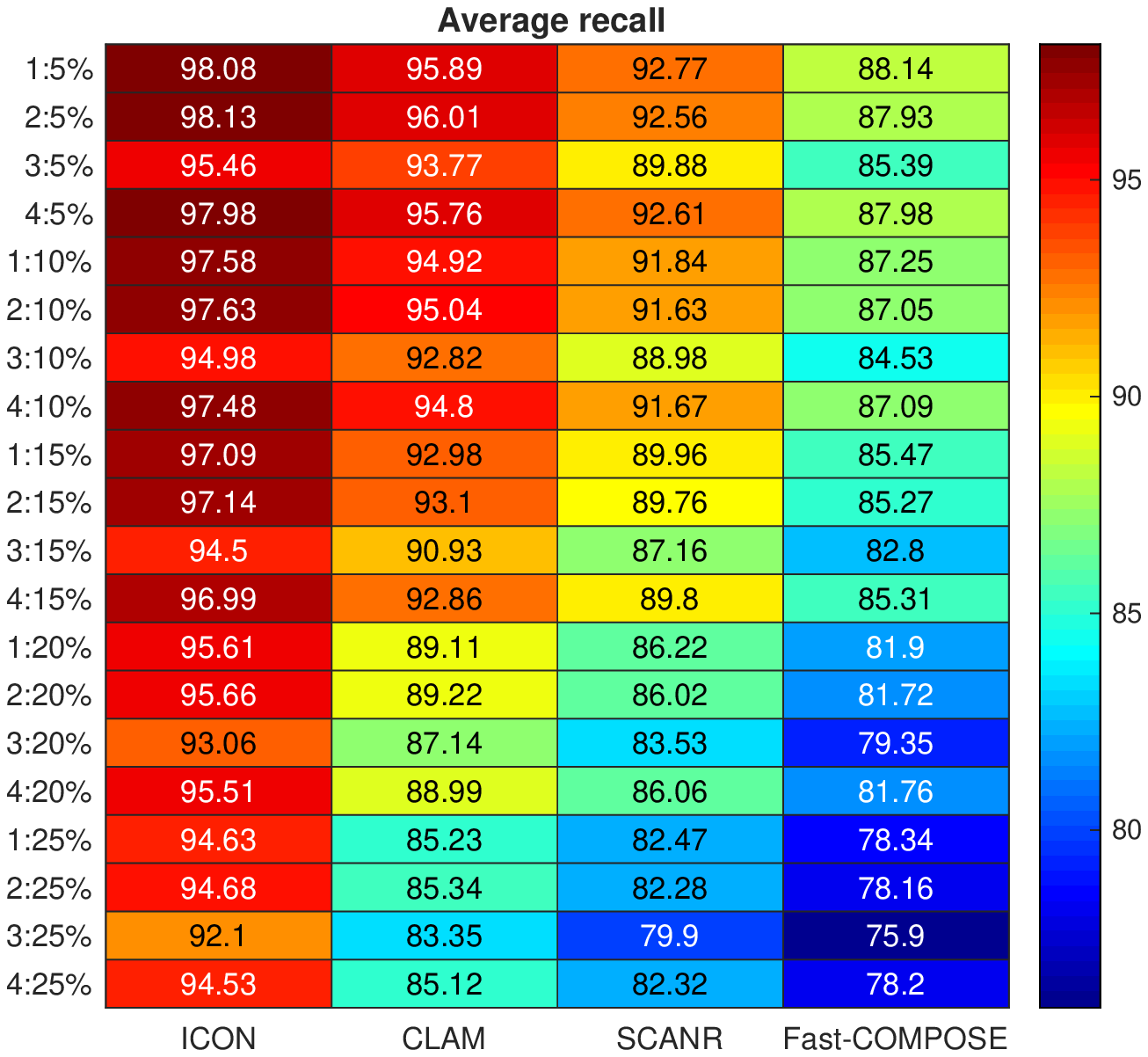}
    \caption{Average recall w.r.t. noise degree $\sigma$. Labels on the Y-axis show the scenario number versus $\sigma$.}
    \label{fig:heat_recall}
\end{figure}

So far, the experiments were based on the CNYPS dataset, which was a collection of real-world PMU signals. In order to further complicate the classification task for the classifiers and the detector, we additionally add independent Gaussian noise $\mathcal{N}(0, \sigma^2)$ to half of the columns in each scenario, where the noise degree $\sigma$ changes from 5 to 25 percents. Fig. \ref{fig:heat_acc} shows the averaged accuracy of classifiers over different scenarios and noise ratios. Changes in the color spectrum of each column indicates the sensitivity of accuracy performance to $\sigma$. As expected, Fig. \ref{fig:heat_acc} shows that the injected noise degrades the performance. However, ICON exhibits superior robustness against the added noise compared to the rest of the data-driven methods. The same pattern can be observed for precision and recall in Fig. \ref{fig:heat_pre} and Fig. \ref{fig:heat_recall}, respectively. Interestingly, the ranking among the selected classifiers is similar to cases without Gaussian noise, and previous conclusions also hold for this analysis.

\subsubsection{Signal Retrieval}

\begin{figure}[t]
\centering
\includegraphics[trim={0.4cm 0.7cm 1.4cm 0.6cm},clip,width=\columnwidth]{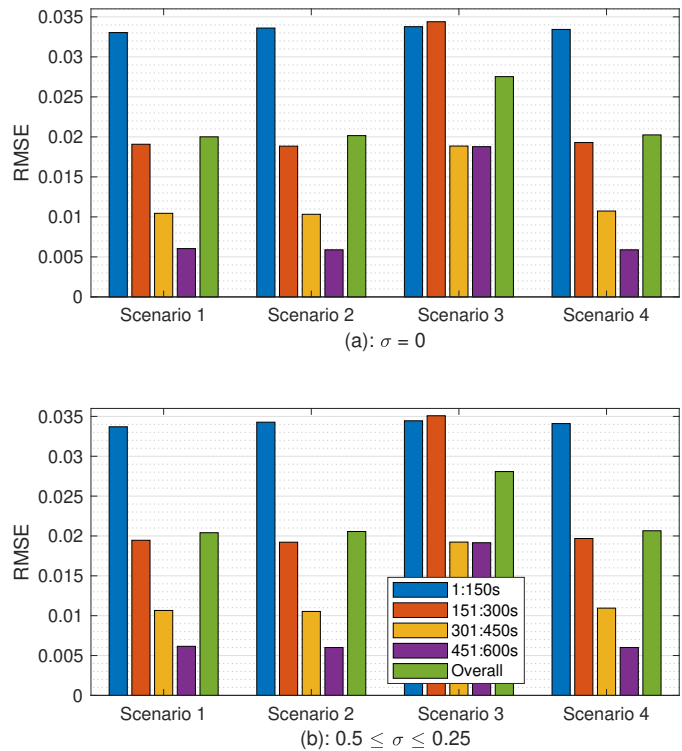}
\caption{Averaged root mean square errors (RMSE) upon signal retrieval for all channels in different time intervals.}
\label{fig:rmse}
\end{figure}

Fig. \ref{fig:signal} compares the outcome of signal retrieval procedure with the original and the disrupted control signals. To evaluate the retrieved signal, we compute the root mean square error (RMSE) for each channel of the PMU data. The averaged RMSE values for each scenario are shown in Fig. \ref{fig:rmse}. The attained results indicate promising performance for signal retrieval over all scenarios. Moreover, the difference between noiseless and noisy cases are negligible.

\begin{figure}[t]
\centering
\includegraphics[trim={0.9cm 0cm 1.24cm 0.4cm},clip,width=\columnwidth]{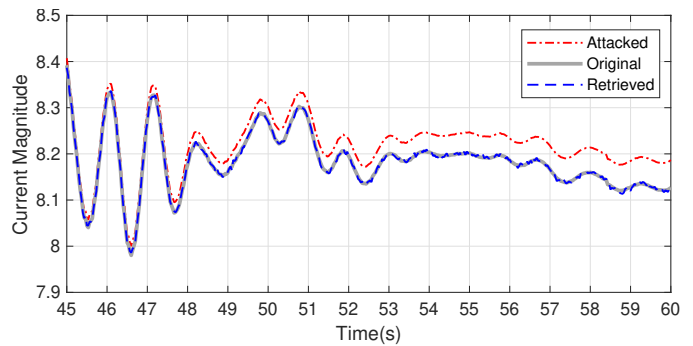}
\caption{Signal retrieval on the second channel of the first scenario within a 15-$s$ time period with $\sigma = 0$.}
\label{fig:sig_rec}
\end{figure}

Apparently, when unobservable FDI attacks change the signal magnitude with a slower rate, they are more difficult to deal with, and vice versa. For instance, within the 1:150$s$ time interval, the unobservable attack slightly changes the signal magnitude in all scenarios (see Fig. \ref{fig:Attack}), whereas changes are more noticeable during the 451:600$s$ interval. Accordingly, the highest and lowest RMSE values are related to the first and last time intervals (see Fig. \ref{fig:signal}), respectively. However, even for the first interval, the retrieved signal is very precise, as shown in Fig. \ref{fig:sig_rec}. Small RMSE values obtained through our experiments imply high accuracy of the control signal retrieval module.

The simulations are performed using MATLAB on a computer with an Intel Core i7-6700HQ CPU and 8Gb of RAM. The average run time of the proposed framework for each scenario is roughly 4.82 seconds. This includes processing 18000 samples of multiple channels through detection, classification, signal retrieval, and self-updates.

\section{Conclusion}
\label{sec:conc}
In this paper, a framework is proposed for the detection and classification of unobservable FDI attacks, which consists of three newly designed modules. The first module enables the detection of unobservable FDI attacks. This module is designed based on sliding least-squares fitting on a sliding window that moves over the PMU signal on a complex plane. Using this approach, one can model historical changes w.r.t. the deviations caused by an attack from the origin on the complex plane and extract attack patterns from the signal. These attack patterns are then used by the second module to perform online semi-supervised classification. The proposed classifier can operate in non-stationary environments and update itself without dependence on external updates. The last module uses the extracted knowledge by other modules to retrieve the control signal before sending it to the control unit. In order to evaluate the designed system, a case study on the CNYPS is given. Additionally, two advanced real-time classifiers are selected for the sake of comparison. The obtained results indicate that the proposed framework is satisfyingly robust, even when FDI attacks are aligned with a disturbance in the system. The detection module is very sensitive to unobservable attacks. The classification module shows superior stability and accuracy over its rivals. The control signal retrieval module results in minimal RMSE.

\ifCLASSOPTIONcaptionsoff
  \newpage
\fi

\section*{Appendix 1}
\label{sec:apx2}
Assume that the PMU measurement vectors at each time instant $t$ can be modeled as:
\begin{equation}
    m_t = Hs_t + \epsilon_t,
\end{equation}
where $m_t$ is a row of $M$; $s_t$ is the $p \times 1$ state vector of complex bus voltages; $\epsilon_t$ is an i.i.d $n \times 1$ additive Gaussian noise vector; $H$ is the measurement Jacobian matrix of size $n\times p$. The least squares estimate of $s_t$, $\hat{s}_t$, can be obtained by:
\begin{equation}
    \hat{s}_t = H^{+} M_t
    \label{eq:x}
\end{equation}
where $(\cdot)^+$ indicates pseudo-inverse operation. The associated residual vector can be defined as:
\begin{equation}
    r = m_t - H\hat{s}_t = m_t - HH^+m_t.
\end{equation}
Conventional bad data detectors (BDD) monitor $r$ to detect bad measurements.

Assuming that the attacker has control over a subset of measurements $J \subseteq \{1, \dots, n\}$, the attacker can manipulate $M$ as:
\begin{equation}
    \Bar{M} = M + D,    
\end{equation}
where the column support of the measurement attack matrix $D$, $\operatorname{Sup}(D)$, is contained in $J$. An attack can bypass the conventional residual-based BDD if $D = CH^T$ for some non- zero matrix $C$:
\begin{equation}
    \Bar{M} = M + CH^T
\end{equation}
From (\ref{eq:x}), the estimated state matrix without attack is given by
\begin{equation}
    \Bar{S} = \Bar{M}H^{+T} + CH^T H^{+T} = \hat{S} + C.
\end{equation}
Therefore, the residual matrix under attack can be obtained as:
\begin{equation}
\begin{split}
    R = \Bar{M} - &\Bar{S}H^T =  \\& M+CH^T - (\hat{S} + C)H^T = M- \hat{S}H^T,
\end{split}
\end{equation}
which is the same as the residual without attack. Thus, attacks in the form above cannot be detected by residue-based BDD.

\section*{Appendix 2}
\label{sec:apx1}
The proof of Proposition \ref{th:th1} is explained in the following.
\begin{proof}
$\forall W_t$ $\wedge$ $e_t>0$, the fitted circle to $z_r\in W_t$, $1\leq r\leq \omega$ has the algebraic representation in the following form:
\begin{equation}\label{eq:eq1}
F(\mathbf{x}+e) = a(\mathbf{x}+e)^T(\mathbf{x}+e) + b^T(\mathbf{x}+e) + c = 0~,
\end{equation}
where $a\neq 0$, and $\mathbf{x}$ is the coordinate vector. The coefficients $\mathbf{u} = (a,b_1,b_2,c)^T$ in Equation (\ref{eq:eq1}) can be found by forming a linear equation system $B\mathbf{u} = 0$, in which $B$ contains the coordinates of data points attained by considering the imaginary and real part of $z_i\in W_t$ as follows:
\begin{equation}\label{eq:eq2}
B_t=\begin{bmatrix}
	x_{1}+ y_{1}+e_{11}+e_{12}& x_{1}+e_{11} & y_{1}+e_{12} &1\\
    \vdots & \vdots & \vdots & \vdots \\
    x_{\omega}+ y_{\omega} + e_{\omega1}+e_{\omega2} & x_{\omega}+e_{\omega1} & y_{\omega}+e_{\omega2} &1\\
\end{bmatrix}
\end{equation}
However, for $\omega>3$, this linear system only has a solution when $z_i\in W_t$ are all located exactly on one circle, which is not always the case. Thus, an optimization problem is used instead to find the best possible fit on $z_i\in W_t$ as follows: 
\begin{equation}\label{eq:eq3}
\min{\parallel B_t \mathbf{u}\parallel}~~~\operatorname{s.t.} \parallel \mathbf{u}\parallel = 1~,
\end{equation}
which is equivalent to finding the right singular vector associated with the smallest singular value of $B_t$. 
Assuming $a\neq 0$, Equation (\ref{eq:eq1}) can be converted to the following form: 
\begin{equation}\label{eq:eq4}
\left(x+e_1 + \frac{b_1}{2a}\right)^2 + \left(y + e_2+ \frac{b_2}{2a}\right)^2 = \frac{\parallel b\parallel^2}{4a^2} - \frac{c}{a}
\end{equation}
\begin{equation}\label{eq:eq5}
\begin{split}
\Longrightarrow\hat{c_t} = &\left(-e_1-\frac{b_1}{2a}, -e_2- \frac{b_2}{2a}\right)\\ = &\left(-\frac{b_1}{2a},-\frac{b_2}{2a}\right) +\left(-e_1,-e_2\right) = \tilde{c}_t + c_t~,
\end{split}
\end{equation}
where, based on Assumption \ref{as:as1}, $\tilde{c}_t$ can be modeled as excessive deviation of $c_t$ from $c_0$, and, therefore, Proposition \ref{th:th1} is proved.
\end{proof}

\section*{Acknowledgment}
This work is supported by the Natural Sciences and Engineering Research Council of Canada (NSERC) under funding reference numbers CGSD3-569341-2022 and RGPIN-2021-02968.

\bibliography{References}

\begin{thebibliography}{10}
\providecommand{\url}[1]{#1}
\csname url@samestyle\endcsname
\providecommand{\newblock}{\relax}
\providecommand{\bibinfo}[2]{#2}
\providecommand{\BIBentrySTDinterwordspacing}{\spaceskip=0pt\relax}
\providecommand{\BIBentryALTinterwordstretchfactor}{4}
\providecommand{\BIBentryALTinterwordspacing}{\spaceskip=\fontdimen2\font plus
\BIBentryALTinterwordstretchfactor\fontdimen3\font minus
  \fontdimen4\font\relax}
\providecommand{\BIBforeignlanguage}[2]{{%
\expandafter\ifx\csname l@#1\endcsname\relax
\typeout{** WARNING: IEEEtran.bst: No hyphenation pattern has been}%
\typeout{** loaded for the language `#1'. Using the pattern for}%
\typeout{** the default language instead.}%
\else
\language=\csname l@#1\endcsname
\fi
#2}}
\providecommand{\BIBdecl}{\relax}
\BIBdecl

\bibitem{9207760}
M.~Higgins, F.~Teng, and T.~Parisini, ``Stealthy mtd against unsupervised
  learning-based blind fdi attacks in power systems,'' \emph{IEEE Trans. Inf.
  Forensics Security}, vol.~16, pp. 1275--1287, 2021.

\bibitem{8660426}
S.~Ahmed, Y.~Lee, S.-H. Hyun, and I.~Koo, ``Unsupervised machine learning-based
  detection of covert data integrity assault in smart grid networks utilizing
  isolation forest,'' \emph{IEEE Trans. Inf. Forensics Security}, vol.~14,
  no.~10, pp. 2765--2777, 2019.

\bibitem{7867825}
R.~Tan, H.~H. Nguyen, E.~Y.~S. Foo, D.~K.~Y. Yau, Z.~Kalbarczyk, R.~K. Iyer,
  and H.~B. Gooi, ``Modeling and mitigating impact of false data injection
  attacks on automatic generation control,'' \emph{IEEE Trans. Inf. Forensics
  Security}, vol.~12, no.~7, pp. 1609--1624, 2017.

\bibitem{7433442}
X.~Liu, Z.~Li, X.~Liu, and Z.~Li, ``Masking transmission line outages via false
  data injection attacks,'' \emph{IEEE Trans. Inf. Forensics Security},
  vol.~11, no.~7, pp. 1592--1602, 2016.

\bibitem{7792204}
Y.~Zhao, A.~Goldsmith, and H.~V. Poor, ``Minimum sparsity of unobservable power
  network attacks,'' \emph{IEEE Trans. Autom. Control}, vol.~62, no.~7, pp.
  3354--3368, July 2017.

\bibitem{7529082}
P.~Gao, M.~Wang, J.~H. Chow, S.~G. Ghiocel, B.~Fardanesh, G.~Stefopoulos, and
  M.~P. Razanousky, ``Identification of successive "unobservable" cyber data
  attacks in power systems through matrix decomposition,'' \emph{IEEE Trans.
  Signal Process.}, vol.~64, no.~21, pp. 5557--5570, Nov 2016.

\bibitem{8472173}
M.~Khalaf, A.~Youssef, and E.~El-Saadany, ``Joint detection and mitigation of
  false data injection attacks in agc systems,'' \emph{IEEE Trans. Smart Grid},
  pp. 1--1, 2018.

\bibitem{8409487}
M.~N. Kurt, Y.~Yılmaz, and X.~Wang, ``Real-time detection of hybrid and
  stealthy cyber-attacks in smart grid,'' \emph{IEEE Trans. Inf. Forensics
  Security}, vol.~14, no.~2, pp. 498--513, 2019.

\bibitem{6032057}
O.~Kosut, L.~Jia, R.~J. Thomas, and L.~Tong, ``Malicious data attacks on the
  smart grid,'' \emph{IEEE Trans. Smart Grid}, vol. 2(4), pp. 645--658, 2011.

\bibitem{4633363}
P.~K. {Mallapragada}, R.~{Jin}, A.~K. {Jain}, and Y.~{Liu}, ``Semiboost:
  Boosting for semi-supervised learning,'' \emph{IEEE Trans. Pattern Anal.
  Mach. Intell.}, vol.~31, no.~11, pp. 2000--2014, 2009.

\bibitem{5444873}
K.~{Chen} and S.~{Wang}, ``Semi-supervised learning via regularized boosting
  working on multiple semi-supervised assumptions,'' \emph{IEEE Trans. Pattern
  Anal. Mach. Intell.}, vol.~33, no.~1, pp. 129--143, 2011.

\bibitem{8628992}
J.~Camacho, G.~Maci{\'a}-Fern{\'a}ndez, N.~M. Fuentes-Garc{\'\i}a, and
  E.~Saccenti, ``Semi-supervised multivariate statistical network monitoring
  for learning security threats,'' \emph{IEEE Trans. Inf. Forensics Security},
  vol.~14, no.~8, pp. 2179--2189, 2019.

\bibitem{6604410}
K.~B. Dyer, R.~Capo, and R.~Polikar, ``Compose: A semisupervised learning
  framework for initially labeled nonstationary streaming data,'' \emph{IEEE
  Trans. Neural Netw.}, vol.~25, no.~1, pp. 12--26, Jan 2014.

\bibitem{5453372}
M.~Masud, J.~Gao, L.~Khan, J.~Han, and B.~M. Thuraisingham, ``Classification
  and novel class detection in concept-drifting data streams under time
  constraints,'' \emph{IEEE Trans. Knowl. Data Eng.}, vol.~23, no.~6, pp.
  859--874, June 2011.

\bibitem{7350165}
T.~Al-Khateeb, M.~M. Masud, K.~M. Al-Naami, S.~E. Seker, A.~M. Mustafa,
  L.~Khan, Z.~Trabelsi, C.~Aggarwal, and J.~Han, ``Recurring and novel class
  detection using class-based ensemble for evolving data stream,'' \emph{IEEE
  Trans. Knowl. Data Eng.}, vol.~28, no.~10, pp. 2752--2764, Oct 2016.

\bibitem{7332782}
Y.~Yuan, D.~Ma, and Q.~Wang, ``Hyperspectral anomaly detection by graph pixel
  selection,'' \emph{IEEE Trans. Cybern.}, vol.~46, no.~12, pp. 3123--3134, Dec
  2016.

\bibitem{7313024}
J.~Zhao, G.~Zhang, M.~L. Scala, Z.~Y. Dong, C.~Chen, and J.~Wang, ``Short-term
  state forecasting-aided method for detection of smart grid general false data
  injection attacks,'' \emph{IEEE Trans. Smart Grid}, vol.~8, no.~4, pp.
  1580--1590, July 2017.

\bibitem{book}
A.~Abur and A.~Gomez-Exposito, \emph{Power System State Estimation: Theory and
  Implementation}, 03 2004, vol.~24.

\bibitem{4074022}
F.~C. Schweppe and J.~Wildes, ``Power system static-state estimation, part i:
  Exact model,'' \emph{IEEE Trans. Power App. Syst.}, vol. PAS-89, no.~1, pp.
  120--125, 1970.

\bibitem{8624411}
J.~Zhao, A.~G{\'o}mez-Exp{\'o}sito, M.~Netto, L.~Mili, A.~Abur, V.~Terzija,
  I.~Kamwa, B.~Pal, A.~K. Singh, J.~Qi, Z.~Huang, and A.~P.~S. Meliopoulos,
  ``Power system dynamic state estimation: Motivations, definitions,
  methodologies, and future work,'' \emph{IEEE Trans. Power Syst.}, vol.~34,
  no.~4, pp. 3188--3198, 2019.

\bibitem{8278264}
M.~N. Kurt, Y.~Yılmaz, and X.~Wang, ``Distributed quickest detection of
  cyber-attacks in smart grid,'' \emph{IEEE Trans. Inf. Forensics Security},
  vol.~13, no.~8, pp. 2015--2030, 2018.

\bibitem{8759957}
------, ``Secure distributed dynamic state estimation in wide-area smart
  grids,'' \emph{IEEE Trans. Inf. Forensics Security}, vol.~15, pp. 800--815,
  2020.

\bibitem{9409791}
M.~Kordestani and M.~Saif, ``Observer-based attack detection and mitigation for
  cyberphysical systems: A review,'' \emph{IEEE Syst., Man, Cybern. Mag.},
  vol.~7, no.~2, pp. 35--60, 2021.

\bibitem{9318019}
M.~Shi, X.~Chen, M.~Shahidehpour, Q.~Zhou, and J.~Wen, ``Observer-based
  resilient integrated distributed control against cyberattacks on sensors and
  actuators in islanded ac microgrids,'' \emph{IEEE Trans. Smart Grid},
  vol.~12, no.~3, pp. 1953--1963, 2021.

\bibitem{9129765}
X.~Luo, Y.~Li, X.~Wang, and X.~Guan, ``Interval observer-based detection and
  localization against false data injection attack in smart grids,'' \emph{IEEE
  Internet Things J.}, vol.~8, no.~2, pp. 657--671, 2021.

\bibitem{8752431}
Z.~Cao, Y.~Niu, and J.~Song, ``Finite-time sliding-mode control of markovian
  jump cyber-physical systems against randomly occurring injection attacks,''
  \emph{IEEE Trans. Autom. Control}, vol.~65, no.~3, pp. 1264--1271, 2020.

\bibitem{9547709}
A.~Cecilia, S.~Sahoo, T.~Dragicevic, R.~Costa-Castello, and F.~Blaabjerg, ``On
  addressing the security and stability issues due to false data injection
  attacks in dc microgrids an adaptive observer approach,'' \emph{IEEE Trans.
  Power Electron.}, pp. 1--1, 2021.

\bibitem{Saif2003}
M.~Saif and Y.~Xiong, \emph{Sliding Mode Observers and Their Application in
  Fault Diagnosis}.\hskip 1em plus 0.5em minus 0.4em\relax Berlin, Heidelberg:
  Springer Berlin Heidelberg, 2003, pp. 1--57.

\bibitem{8957680}
X.~Wang, X.~Luo, M.~Zhang, Z.~Jiang, and X.~Guan, ``Detection and isolation of
  false data injection attacks in smart grid via unknown input interval
  observer,'' \emph{IEEE Internet Things J.}, vol.~7, no.~4, pp. 3214--3229,
  2020.

\bibitem{article}
V.~Utkin and H.-C. Chang, ``Sliding mode control on electro-mechanical
  systems,'' \emph{Math. Problems in Eng.}, vol.~8, 09 2002.

\bibitem{7801908}
G.~Chen, Y.~Song, and Y.~Guan, ``Terminal sliding mode-based consensus tracking
  control for networked uncertain mechanical systems on digraphs,'' \emph{IEEE
  Trans. Neural Netw. Learn. Syst}, vol.~29, no.~3, pp. 749--756, 2018.

\bibitem{9055170}
T.~Wu, W.~Xue, H.~Wang, C.~Y. Chung, G.~Wang, J.~Peng, and Q.~Yang, ``Extreme
  learning machine-based state reconstruction for automatic attack filtering in
  cyber physical power system,'' \emph{IEEE Trans. Ind. Informat.}, vol.~17,
  no.~3, pp. 1892--1904, 2021.

\bibitem{8519325}
Z.~Guo, D.~Shi, D.~E. Quevedo, and L.~Shi, ``Secure state estimation against
  integrity attacks: A gaussian mixture model approach,'' \emph{IEEE Trans.
  Signal Process.}, vol.~67, no.~1, pp. 194--207, 2019.

\bibitem{LI2020101705}
Y.~Li and Y.~Wang, ``Developing graphical detection techniques for maintaining
  state estimation integrity against false data injection attack in integrated
  electric cyber-physical system,'' \emph{J. Syst. Architecture}, vol. 105, p.
  101705, 2020.

\bibitem{8336956}
R.~Razavi-Far, E.~Hallaji, M.~Saif, and G.~Ditzler, ``A novelty detector and
  extreme verification latency model for nonstationary environments,''
  \emph{IEEE Trans. Ind. Electron.}, vol.~66, no.~1, pp. 561--570, Jan 2019.

\bibitem{7762928}
N.~Nissim, A.~Cohen, and Y.~Elovici, ``Aldocx: Detection of unknown malicious
  microsoft office documents using designated active learning methods based on
  new structural feature extraction methodology,'' \emph{IEEE Trans. Inf.
  Forensics Security}, vol.~12, no.~3, pp. 631--646, 2017.

\bibitem{6137334}
M.~M. Masud, T.~M. Al-Khateeb, L.~Khan, C.~Aggarwal, J.~Gao, J.~Han, and
  B.~Thuraisingham, ``Detecting recurring and novel classes in concept-drifting
  data streams,'' in \emph{IEEE 11th Int. Conf. Data Mining}, Dec 2011, pp.
  1176--1181.

\bibitem{8587555}
Z.~Chu, J.~Zhang, O.~Kosut, and L.~Sankar, ``Unobservable false data injection
  attacks against pmus: Feasible conditions and multiplicative attacks,'' in
  \emph{IEEE Int. Conf. Commun., Control, and Computing Technologies for Smart
  Grids}, 2018, pp. 1--6.

\end{thebibliography}
\bibliographystyle{IEEEtran}

\renewenvironment{IEEEbiography}[1]
  {\IEEEbiographynophoto{#1}}
  {\endIEEEbiographynophoto}

\vspace{1cm}
\begin{IEEEbiography}
{Ehsan Hallaji} (Graduate Student Member, IEEE) received the B.Sc. degree in software engineering, in 2015, and the M.A.Sc. degree in electrical and computer engineering from the University of Windsor, Windsor, ON, Canada, in 2018. He is currently a Ph.D. candidate at the Department of Electrical and Computer Engineering, University of Windsor. His current research interests include machine learning, data mining, and cybersecurity.
\end{IEEEbiography}

\begin{IEEEbiography}
{Roozbeh Razavi-Far} (Senior Member, IEEE) received the B.Sc. degree in Electrical and Computer Engineering, the M.Sc. and Ph.D. degrees from the Amirkabir University of Technology and achieved a second Ph.D. degree from Politecnico di Milano. He is currently an Assistant Professor with the Faculty of Computer Science, University of New Brunswick and an Adjunct Professor with the Department of Electrical and Computer Engineering, University of Windsor. His research focuses on machine learning, computational intelligence, big data analytics, cybernetics, and cybersecurity.
\end{IEEEbiography}

\begin{IEEEbiography}
{Mehrdad Saif} (Senior Member, IEEE) received the B.S., M.S., and D.Eng. degrees in electrical engineering from Cleveland State University, Cleveland, OH, USA, in 1982, 1984, and 1987, respectively. Currently, he is a Professor with the Department of Electrical and Computer Engineering Department, University of Windsor. His research interests include control systems, cybersecurity, condition monitoring, applied machine learning, and and their application in automotive, power, and autonomous systems.
\end{IEEEbiography}

\begin{IEEEbiography}
{Meng Wang} (Member, IEEE) received B.S. and M.S. degrees from Tsinghua University, China, in 2005 and 2007, respectively and the Ph.D. degree from Cornell University, Ithaca, NY, USA, in 2012. She is currently an Associate Professor with the Department of Electrical, Computer, and Systems Engineering, Rensselaer Polytechnic Institute, Troy, NY, USA. Her research interests include high- dimensional data analytics, machine learning, power systems monitoring, and synchrophasor technologies.
\end{IEEEbiography}

\begin{IEEEbiography}
{Bruce Fardanesh} (Fellow, IEEE) received his Doctor of Engineering degree in electrical engineering from Cleveland State University, Cleveland, OH, USA, in 1985. He joined New York Power Authority in 1991, where he is currently the Chief Electrical Engineer. His research interests include power system analysis, modeling, dynamics, operation, and control.
\end{IEEEbiography}

\end{document}